\documentclass[letterpaper, 10 pt, conference]{ieeeconf}  % Comment this line out if you need a4paper

\IEEEoverridecommandlockouts                              % This command is only needed if
\usepackage{graphicx}
\usepackage{amsmath}
\usepackage{amssymb}
\usepackage{amsfonts}
\usepackage{bm}
\usepackage{silence}
\usepackage{subcaption}
\usepackage{bbm}
\usepackage{epstopdf}
\usepackage{balance}
\usepackage{color}
\usepackage{booktabs, array}
\usepackage{xcolor}
\usepackage{makeidx}
\usepackage[toc]{glossaries}
\newacronym{RL}{RL}{reinforcement learning}
\newacronym{DRL}{DRL}{deep reinforcement learning}
\newglossaryentry{DNN}
{
    name=DNN,
    plural=DNNs,
    description={deep neural network}
}
\newacronym{MCTS}{MCTS}{Monte Carlo tree search}
\newacronym{LSTM}{LSTM}{long short-term memory}
\newacronym{DP}{DP}{dynamic programming}
\newacronym{SVM}{SVM}{support vector machine}
\newacronym{SLAM}{SLAM}{simultaneous localization and mapping problem}
\newacronym{MDP}{MDP}{Markov decision process}
\newacronym{POMDP}{POMDP}{partially observable Markov decision process}
\newacronym{AI}{AI}{artificial intelligence}
\newacronym{TD}{TD}{temporal difference}
\newacronym{MSE}{MSE}{mean squared error}
\newacronym{DQN}{DQN}{deep Q networks}
\newacronym{DDPG}{DDPG}{deep deterministic policy gradient}
\newacronym{DPG}{DPG}{deterministic policy gradient}
\newacronym{A3C}{A3C}{asynchronous advantage actor-critic}
\newacronym{A2C}{A2C}{advantage actor-critic}
\newacronym{ReLU}{ReLU}{rectified linear unit}

\newcommand{\url}[1]{#1}

\newenvironment{customBoxed}
 {\trivlist\nopagebreak
  \parindent0pt
  \item\relax\obeylines}
 {\par
  \nopagebreak
  \vspace{0.2em}%
  \endtrivlist}

\def\dmax{\mbox{dmax}}
\def\dist{\mbox{dist}}

\newcommand{\figref}[1]{Fig.~\ref{#1}}

\title{\LARGE \bf Vision-based Navigation Using Deep Reinforcement Learning}

\author{Jon\'a\v{s} Kulh\'anek$^{1}$, Erik Derner$^{2}$, Tim de Bruin$^{1}$, and Robert Babu\v{s}ka$^{3}$% <-this % stops a space
\thanks{$^{1}$Jon\'a\v{s} Kulh\'anek and Tim de Bruin are with Cognitive Robotics, Faculty of 3mE, Delft University of Technology, 2628 CD Delft, The Netherlands {\tt\footnotesize jonas.kulhanek@live.com}, {\tt\footnotesize t.d.debruin@tudelft.nl}}%
\thanks{$^{2}$Erik Derner is with the Czech Institute of Informatics, Robotics and Cybernetics, Czech Technical University in Prague, 16636 Prague, Czech Republic and with the Department of Control Engineering, Faculty of Electrical Engineering, Czech Technical University in Prague, 16627 Prague, Czech Republic {\tt\footnotesize erik.derner@cvut.cz}}%
\thanks{$^{3}$Robert Babu\v{s}ka is with Cognitive Robotics, Faculty of 3mE, Delft University of Technology, 2628 CD Delft, The Netherlands and with the Czech Institute of Informatics, Robotics and Cybernetics, Czech Technical University in Prague, 16636 Prague, Czech Republic {\tt\footnotesize r.babuska@tudelft.nl}}%
\thanks{This work was supported by the European Regional Development Fund under the project IMPACT (reg. no. CZ.02.1.01/0.0/0.0/15\_003/0000468) and under the project Robotics for Industry 4.0 (reg. no. CZ.02.1.01/0.0/0.0/15\_003/0000470).}%
}%

\IEEEoverridecommandlockouts
\pubid{\raisebox{-3\baselineskip}{\begin{minipage}{\textwidth}\
\copyright2019 IEEE. Personal use of this material is permitted. Permission from IEEE must be obtained for all other uses, in any current or future media, including reprinting/republishing this material for advertising or promotional purposes, creating new collective works, for resale or redistribution to servers or lists, or reuse of any copyrighted component of this work in other works. \\
DOI: 10.1109/ECMR.2019.8870964
\end{minipage}}}

\begin{document}

\maketitle
\pagestyle{empty}

%%%%%%%%%%%%%%%%%%%%%%%%%%%%%%%%%%%%%%%%%%%%%%%%%%%%%%%%%%%%%%%%%%%%%%%%%%%%%%%%
\begin{abstract}
Deep reinforcement learning (RL) has been successfully applied to a variety of game-like environments. However, the application of deep RL to visual navigation with realistic environments is a challenging task. We propose a novel learning architecture capable of navigating an agent, e.g. a mobile robot, to a target given by an image. To achieve this, we have extended the batched A2C algorithm with auxiliary tasks designed to improve visual navigation performance. We propose three additional auxiliary tasks: predicting the segmentation of the observation image and of the target image and predicting the depth-map. These tasks enable the use of supervised learning to pre-train a large part of the network and to reduce the number of training steps substantially.
The training performance has been further improved by increasing the environment complexity gradually over time. An efficient neural network structure is proposed, which is capable of learning for multiple targets in multiple environments. Our method navigates in continuous state spaces and on the AI2-THOR environment simulator outperforms state-of-the-art goal-oriented visual navigation methods from the literature.
\end{abstract}

\begin{keywords}
Robot navigation, deep reinforcement learning, actor-critic, auxiliary tasks.
\end{keywords}
%%%%%%%%%%%%%%%%%%%%%%%%%%%%%%%%%%%%%%%%%%%%%%%%%%%%%%%%%%%%%%%%%%%%%%%%%%%%%%%%

%\tableofcontents

%TODO:cite
%Sascha Lange and Martin Riedmiller. Deep auto-encoder neural networks in reinforcement
%learning. In Neural Networks (IJCNN), The 2010 International Joint Conference on, pages
%1–8. IEEE, 2010.
%

\section{Introduction}
\pubidadjcol
\enlargethispage{2\baselineskip}

Visual navigation is the problem of navigating an agent, e.g. a mobile robot, in an environment using camera input only. The agent is given a target image (an image it will see from the target position), and its goal is to move from its current position to the target by applying a sequence of actions, based on the camera observations only. We focus on the case when the environment is initially unknown, i.e., no explicit map is available. Such a visual navigation problem can be formalized as a reinforcement learning (RL) problem \cite{Sutton:1998:IRL:551283}.
Two main challenges in the RL formulation are the dimensionality of the agent's observation space and the fact that the actual state is only partially observable from the images.
%\JK{is this really our state? perhaps we need also the map and the position of the goal to be in the state if the method is to be able to generalize. do we really know what the real state is?}(position and orientation of the agent)

Observation space dimensionality can be reduced by using hand-crafted features, or by using features learned on either the training dataset or a completely different dataset, e.g., ResNet \cite{He2016DeepRL} features automatically extracted from the image \cite{7989381, bruce2017oneshot}. A different method, proposed in \cite{wu2018building}, uses image segmentation and a depth map as inputs to the agent. It was trained and evaluated on houses from SUNCG dataset \cite{song2016ssc}, and the trained agent was able to find multiple targets specified as a separate input to the agent.

Raw high-dimensional input images can also be used directly for navigation \cite{jaderberg2016reinforcement, mirowski2016learning}. These two papers extend the \gls{A3C} algorithm with auxiliary tasks to stabilize the training and make it more efficient when the reward is sparse. They, however, use the DeepMind Lab \cite{beattie2016deepmind} game simulator, which is much simpler than realistic simulators \cite{ai2thor,wu2018building,song2016ssc}. The only method that relies solely on visual input in a realistic indoor-scene environment is \cite{7989381}. However, it was applied to AI2-THOR \cite{ai2thor} which contains small single-room environments and the action space discretized the environment into a simple grid world.

%JK{ED:explain difference between A2C and A3C}
In our approach, the agent learns to navigate based on the observed raw images only, as opposed to \cite{7989381}, which uses ResNet features. The learning algorithm is based on the batched version of \gls{A2C} \cite{Wu2017ScalableTM}, extended with auxiliary tasks to help the agent to learn useful features also in the absence of informative rewards. During the training of the deep neural network, we use depth-maps and image segmentations as training targets for the auxiliary tasks. In addition, we propose a method to pre-train the neural network before the reinforcement learning algorithm is applied. This is accomplished by transfer learning from one environment to another, gradually increasing the environment complexity. Finally, to address the partial observability problem, we propose a novel neural network architecture that is both efficient and compact. We evaluate our method in realistic indoor-scene environments similar to \cite{7989381} and \cite{mirowski2016learning}.

%end of related work

%The input to the agent is a single RGB image taken from the environment at the agent's position in the direction the agent was facing. Furthermore, an image of the goal is also given to the agent in the form of an RGB image.

%We use a different set of actions for each environment. This enables us to compare our method to other methods which use a specific set of actions.
\pubidadjcol

\section{Preliminaries}
\subsection{Formal setting}
\pubidadjcol
\enlargethispage{2\baselineskip}
The visual navigation problem is a \gls{POMDP}. For example, when the agent faces a wall, there are many states yielding the same or a very similar image.
However, for the ease of notation, we will first introduce the problem as an instance of a standard MDP, using the state $s_t \in \mathcal{S}$ as if it was available to the agent. Later, we will replace the state by a sequence of past observations $o_1, o_2, \ldots, o_t$.

At the beginning of each learning episode, the agent starts from state $s_0$ which is uniformly sampled from the set of all possible initial states $\mathcal{S}_{start}$: $s_0 \sim \mathbb{U}(\mathcal{S}_{start})$.\footnote{Other distributions than the uniform one can be used.} At discrete time steps $t=0,1,2,\ldots$ the agent executes actions $a_t$. As a result of each action, the agent moves to the next state $s_{t+1}$ and receives reward $r_{t+1}$. The experience the agent collects in a single episode is defined as the following sequence:
\begin{equation}\label{eq:zeta_complete}
    \zeta =  s_0, a_0, r_{1}, s_{1}, a_{1}, r_{2}, \ldots \,.
\end{equation}
An episode terminates when the agent reaches the target or after a predefined maximum number of time steps. For training, the episode is split into equally long {\em rollouts}, where the last rollout can be shorter. The experience collected in a single rollout of length $\ell$ is defined as:
\begin{equation}\label{eq:zeta_partial}
    \zeta_t^\ell = s_t, a_t, r_{t+1}, s_{t+1}, a_{t+1}, r_{t+2}, ..., r_{t+\ell+1}, s_{t+\ell+1} \, .
\end{equation}

\subsection{Advantage Actor-Critic Algorithms (A2C)}
Actor-critic algorithms are suitable for continuous state spaces \cite{Grondman12-SMC-survey}. The critic is an approximator of the state-value function: $v_{\theta_v}\colon \mathcal{S} \rightarrow \mathbb{R}$, parameterized by $\theta_v$, while the actor is an approximator of the policy. We use a stochastic policy $\pi_{\theta_p}(a|s)$ which is a probability distribution over the discrete set of possible actions, conditioned on the state $s \in \mathcal{S}$, and parameterized by $\theta_p$.
Let the bootstrapped $n$-step return $g_t^n$ be defined as:
\begin{equation}\label{eq:a2c_reward}
     g_t^n = r_{t+1} + \gamma r_{t+2} + \ldots + \gamma^{n - 1} r_{t + n} + \gamma^{n} \mathbbm{1}_{\textrm{c}} v_{\theta_v}(s_{t + n})\, ,
\end{equation}
where $\mathbbm{1}_{\textrm{c}}$ is zero if the episode ended during the rollout and one otherwise and $n\geq 1$. The actor is updated similarly to REINFORCE  \cite{Williams1992} with advantage estimates from the critic. The gradient of the actor's loss function $J_p$  from \cite{sutton2000policy} is given by:
\begin{equation}\label{eq:a2c_policy_gradient}
     \frac{\partial J_p}{\partial \theta_p} = -\sum_{i = t}^{t + \ell} \frac{\partial}{\partial \theta_p} \log \pi_{\theta_p}(a_{i}|s_{i}) \left(g_{i}^{t + \ell - i + 1} - v_{\theta_v}(s_{i})\right) \, .
\end{equation}
The term $ g_{i}^{t + \ell - i + 1} - v_{\theta_v}(s_{i})$ is referred to as the advantage function. The critic is updated using the $n$-step temporal difference learning: the \Gls{MSE} between the bootstrapped $n$-step return and the critic output is computed and a gradient descent update is applied. The gradient of the critic's loss function $J_v$ is:
\begin{equation}
    \label{eq:value_gradient}
    \frac{\partial J_v}{\partial \theta_v} = \sum_{i = t}^{t + \ell} \frac{\partial}{\partial \theta_v} \frac{1}{2} \bigl(g_i^{t + \ell - i + 1} - v_{\theta_v}(s_i)\bigr)^2 \, .
\end{equation}
%
%We usually combine both gradients into a single gradient descent update.
%The algorithm is on-policy.
To ensure exploration, the negative entropy of the actor is added to the total loss. This negative entropy in state $s$ is defined as:
\begin{equation}
    H^{-}(s, \theta_p) = \sum_{a \in \mathcal{A}} \pi_{\theta_p}(a|s) \log \pi_{\theta_p}(a|s)
\end{equation}
and its gradient on the rollout data is:
\begin{equation}
    \frac{\partial J_e}{\partial \theta_p} = \sum_{i = t}^{t + \ell}\sum_{a \in \mathcal{A}} \frac{\partial}{\partial \theta_p} \pi_{\theta_p}(a|s_{i}) \log \pi_{\theta_p}(a|s_{i})\,.
\end{equation}
Note that the above setting differs from the one given in \cite{Sutton:1998:IRL:551283}, which uses the $n$-step forward view to compute the return $g$. When DNNs are used to approximate the actor and  critic, it is beneficial to optimize on multiple time-steps in a single batch. We, therefore, use the rollout data to optimize all time-steps in the rollout at once. The estimated returns are a mixture of returns with different length for each state, which was proven to reduce the error in the discrete RL setting \cite{Watkins92q-learning,gurvits1994incremental}.

As the critic and actor can share knowledge about the environment, they can share some of their parameters,
which leads to improved learning performance. For example, when using neural networks for visual tasks, the bottom-most convolutional layers in the actor and in the critic need to learn the same convolutional filters. The A2C algorithm \cite{mnih2016asynchronous} has been adapted for the use with DNNs by introducing the following two modifications:
\begin{enumerate}
\item {\em Batched Advantage Actor-Critic (A2C).}
In batched \gls{A2C} \cite{Wu2017ScalableTM}, there are $k$ different environments. At each time step, $k$ actions are sampled by the actor, one for each environment. The rollouts collected from the environments are used to optimize the actor and the critic in a single batch. This process can be viewed as having $k$ separate instances of \gls{A2C}, each updating the same shared parameters. As shown in \cite{mnih2016asynchronous}, the use of multiple environments has a stabilizing effect on the training, similarly to using an experience buffer \cite{mnih2013playing}.
\item {\em Off-policy Critic Updates.}
Collecting observations can be costly, especially when the environment framework has to simulate physics and render 3D scenes. For an algorithm to be efficient, it needs to learn as much as possible from the experiences collected so far. To improve the data efficiency and the stability of the algorithm, a memory of past experiences called the experience buffer is used. It keeps the last $n_e$ experiences, i.e., observations, actions, rewards, and terminals\footnote{A terminal is the indicator of the episode ending in a particular time step.}. At each learning step, a sequence of experiences is sampled from the buffer and it is used to compute the bootstrapped $n$-step returns \eqref{eq:a2c_reward} and so to train the critic.
\end{enumerate}

\subsection{UNREAL Auxiliary Tasks}
Deep RL algorithms are commonly enhanced with auxiliary tasks to improve their learning performance. For instance, in \cite{jaderberg2016reinforcement} the A3C algorithm was extended with two auxiliary tasks, {\em reward prediction}, and {\em pixel control}. The former predicts the sign of the reward based on past four observations and the latter uses an additional pseudo-reward function to learn a policy that maximizes the absolute pixel change. The batched A2C can be enhanced in the same way; more details are given in Section~\ref{sec:aux_rl}.

\section{Proposed Learning Architecture}
Our method extends the batched A2C algorithm with UNREAL auxiliary tasks and additional auxiliary tasks for visual navigation. We call the method A2CAT-VN, which is an abbreviation of A2C with Auxiliary Tasks for Visual Navigation. We have made its implementation\footnote{\url{https://github.com/jkulhanek/a2cat-vn-pytorch}} as well as a framework implementing several deep RL algorithms\footnote{\url{https://github.com/jkulhanek/deep-rl-pytorch}} publicly available on GitHub.

\subsection{Neural Network}
\begin{figure}[htbp]
    \hspace*{-.3cm}\includegraphics[width=1.08\linewidth]{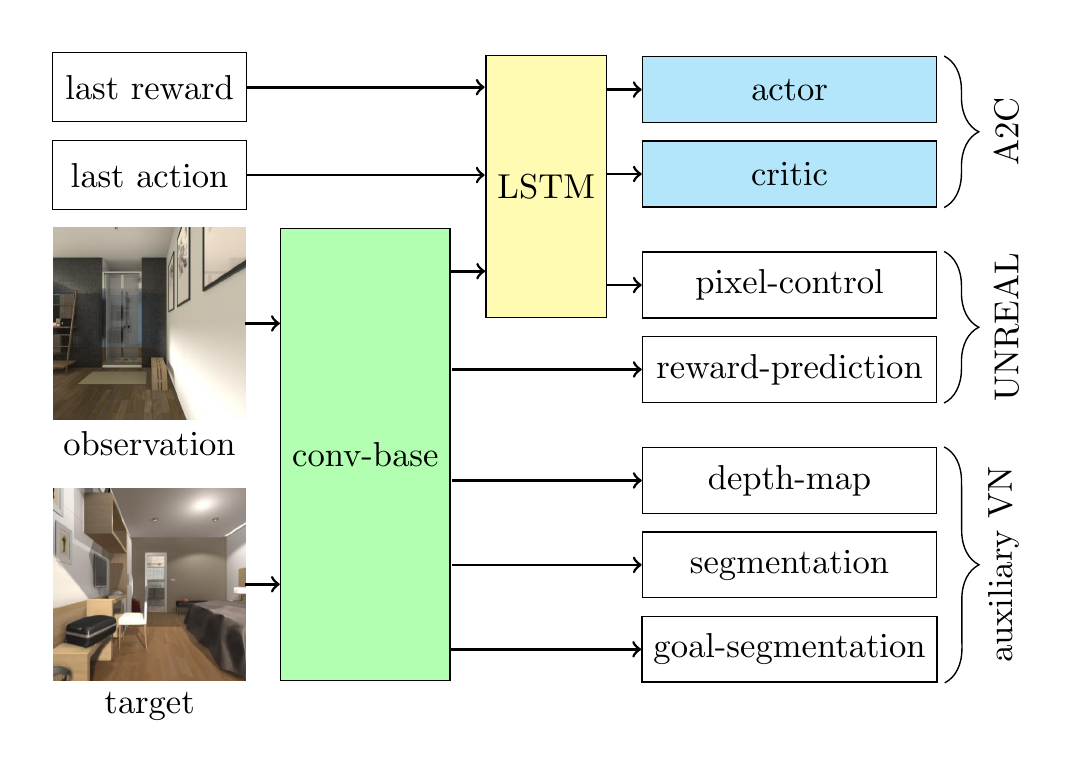}
    \caption{A2CAT-VN neural network architecture.}
    \label{fig:model_overview}
\end{figure}

The deep neural network used in our method consists of several modules: convolutional base, LSTM, actor, critic, and auxiliary tasks, see \figref{fig:model_overview}.
In the sequel, we explain the individual blocks one by one.%

\begin{figure}[htbp]
    \centering
    \includegraphics[width=0.95\linewidth]{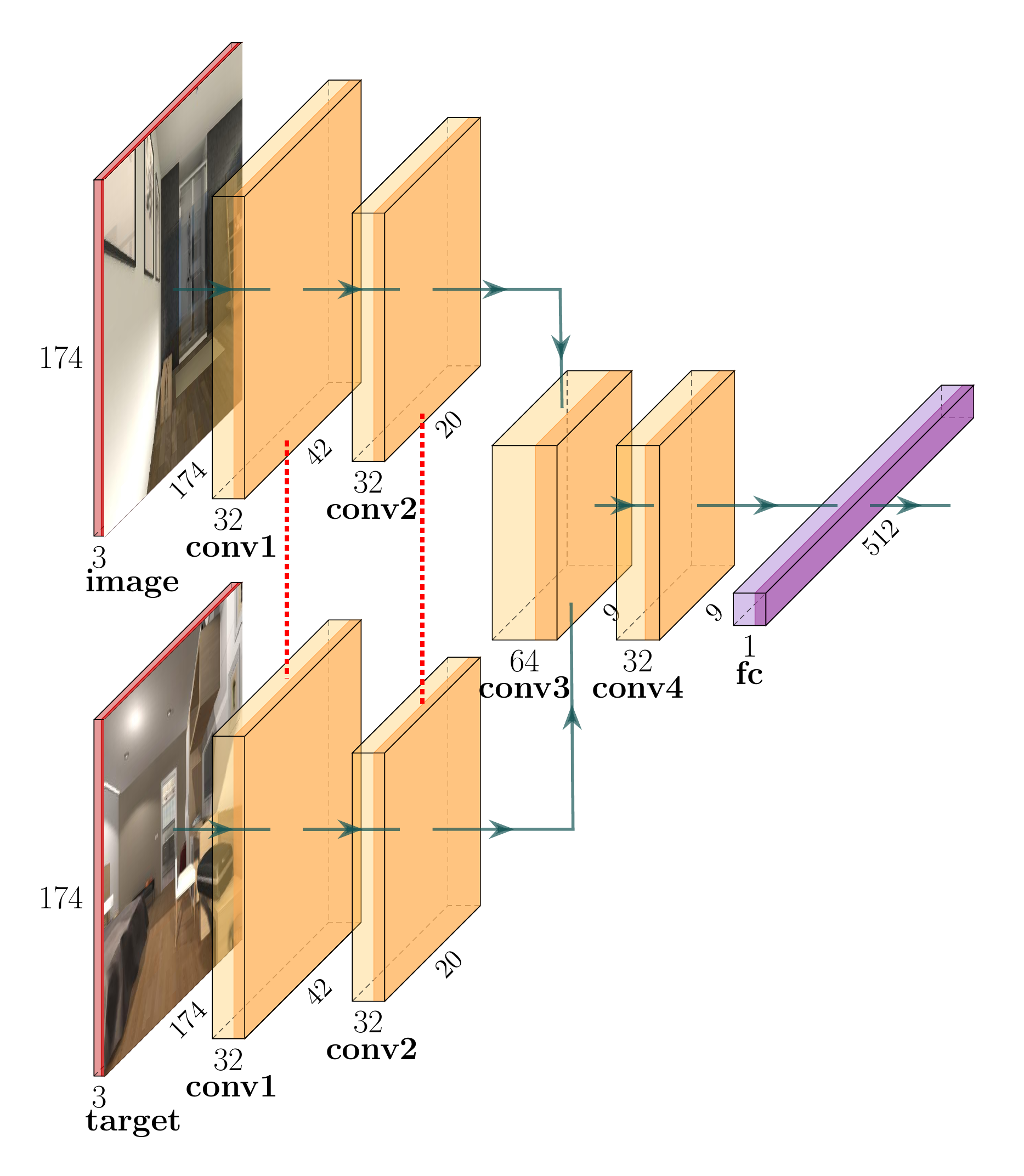}
    \caption{Convolutional base. The feature size after applying each layer is shown in the picture.}
    \label{fig:conv_base}
\end{figure}

The convolutional base is depicted in \figref{fig:conv_base}. Its inputs are the observed image and the target image, each entering into a separate stream of two convolutional layers with shared weight parameters. The outputs of the second layer are concatenated and passed to two additional convolutional layers, followed by a single fully-connected linear layer.

Each of these layers is followed by the ReLU activation function.
We do not employ maxpool layers \cite{He2016DeepRL}. Instead, the images are down-sampled by using stride only as suggested in \cite{springenberg2014striving}.
% It was argued in \cite{springenberg2014striving} that having maxpool layers does not give any performance benefits over strided convolution when applied to vision tasks.
The convolutional base features are merged with the previous action and the previous reward and are passed to the \gls{LSTM} layer \cite{doi:10.1162/neco.1997.9.8.1735}.

The previous action is encoded using one-hot encoding and the previous reward is clipped to the interval $[-1, 1]$. LSTM features are used as the input for both the actor and the critic, as well as for the pixel control auxiliary task. Let $\phi(x)$ be the LSTM features of an input $x$ (LSTM features are computed from the convolutional features and therefore are a function of the input). Note that the input is composed of the image observation, the target image, the last action, and the last reward, as well as the previous LSTM state. The critic is an affine transformation of the LSTM features, and the actor is the result of the softmax function applied to an affine transformation of the LSTM features.

\subsection{Resolving Partial Observability}
The partial observability of the environment does not allow the agent to uniquely distinguish which state it occupies based on a sole observation. Using previous observations can, however, greatly improve its ability to navigate in the environment. For example, if the agent faces a wall, it can instead look at the previous observation and the action taken. The authors of  \cite{mnih2013playing} and \cite{7989381} used the past four frames, fed into the network instead of single image input. In \cite{jaderberg2016reinforcement}, an \gls{LSTM} memory \cite{doi:10.1162/neco.1997.9.8.1735} was used instead. We have used the latter in our approach, as we have experimentally found that it was superior to using the past four frames. Past four frames were not enough to capture the complex experience the agent collected when exploring the environment, and more frames lead to an unmanageable increase of the parameter space size and memory requirements.

\subsection{UNREAL Auxiliary Tasks}
\label{sec:aux_rl}
\subsubsection{Reward Prediction}\label{sec:rp}
The goal of the agent is to maximize the cumulative reward. It proves beneficial to train the network to predict whether a given state leads to a positive reward or not since it helps the network to build useful features to recognize potentially fruitful states. The agent learns to predict the next reward based on the past three observations \cite{jaderberg2016reinforcement, rewardprediction}\footnote{Also here LSTM could be employed. However, we prefer to use the original method from the literature.}. First, a sequence of experiences is sampled from the experience replay buffer such that there is a fixed ratio between the sequences ending with zero reward and the sequences ending with non-zero reward. The output of the fourth convolutional layer computed from all three past observations is merged into a single vector. An additional linear layer and the softmax function are applied to output probabilities of the reward being positive, negative, or zero. This new network is then trained using the cross-entropy loss.

\subsubsection{Pixel Control}\label{sec:pc}
The pixel control task is defined via an additional pseudo-reward function in order to maximize the absolute pixel change. Using this reward, an additional policy is trained that shares most of its parameters with the A2C actor and critic. This policy must be trained using an off-policy RL algorithm since it uses the data sampled from the experience replay buffer generated by the actor. In \cite{jaderberg2016reinforcement} the $n$-step Q-learning loss \cite{mnih2013playing} is used to update the policy. The observation images are downsized, converted to gray scale, and the absolute differences between two consecutive observations are computed and used as pseudo-rewards for Q-learning \cite{mnih2016asynchronous}.

\begin{figure}[htbp]
    \centering
    \includegraphics[width=\linewidth]{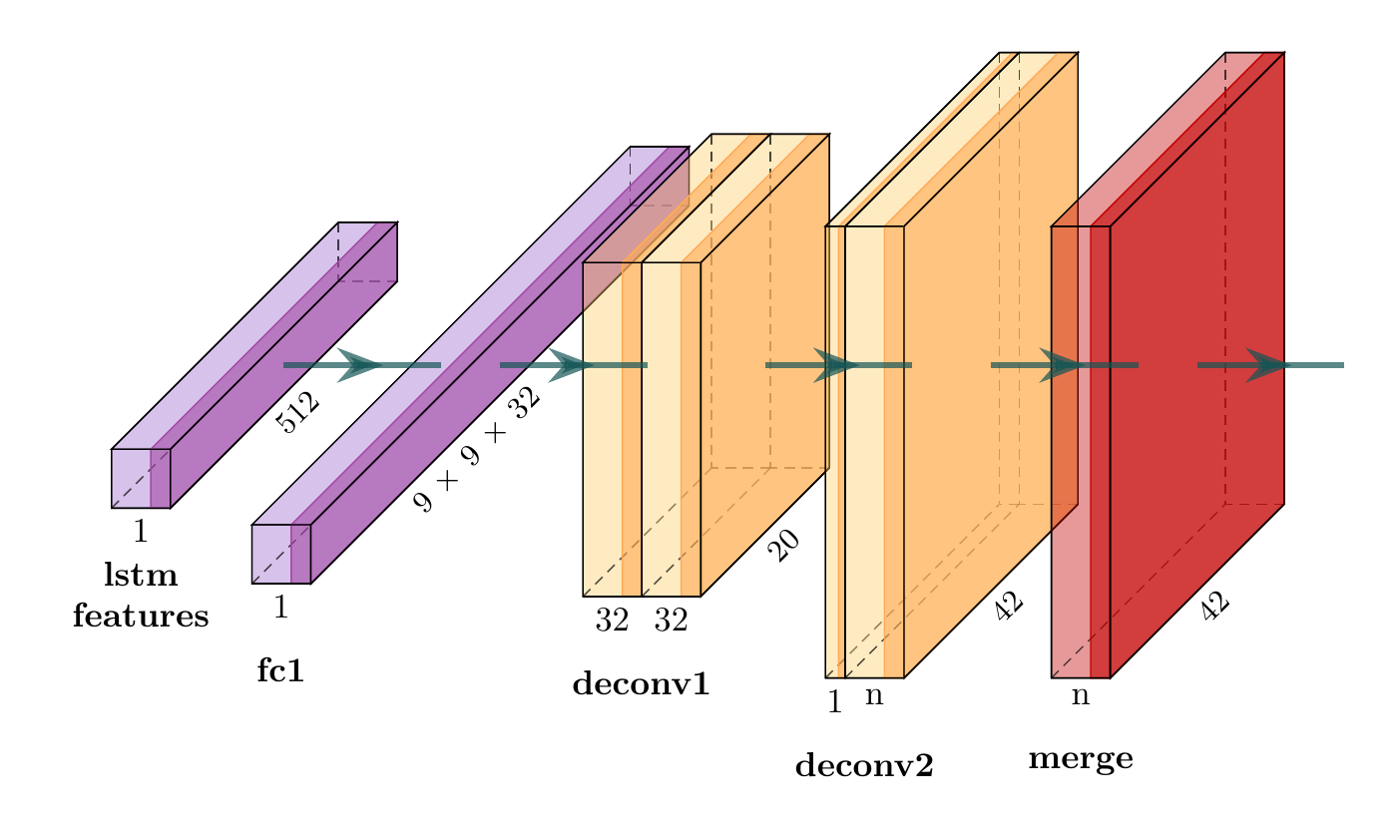}
    \caption{Pixel control network.}
    \label{fig:pc_action}
\end{figure}

A new head is attached to the output of LSTM. This head consists of deconvolutional layers -- upsampling the low-dimensional features back to the size of the downsampled observations. For each action, there is a different output in the last layer to output the Q-function for each pixel. The dueling DQN technique \cite{wang2015dueling} is used to improve the performance of the pixel control network. The pixel control network used in our method can be seen in \figref{fig:pc_action}.

\subsection{Additional Auxiliary Tasks for Visual Navigation}
\begin{figure}
    \centering
    \includegraphics[width=0.90\linewidth]{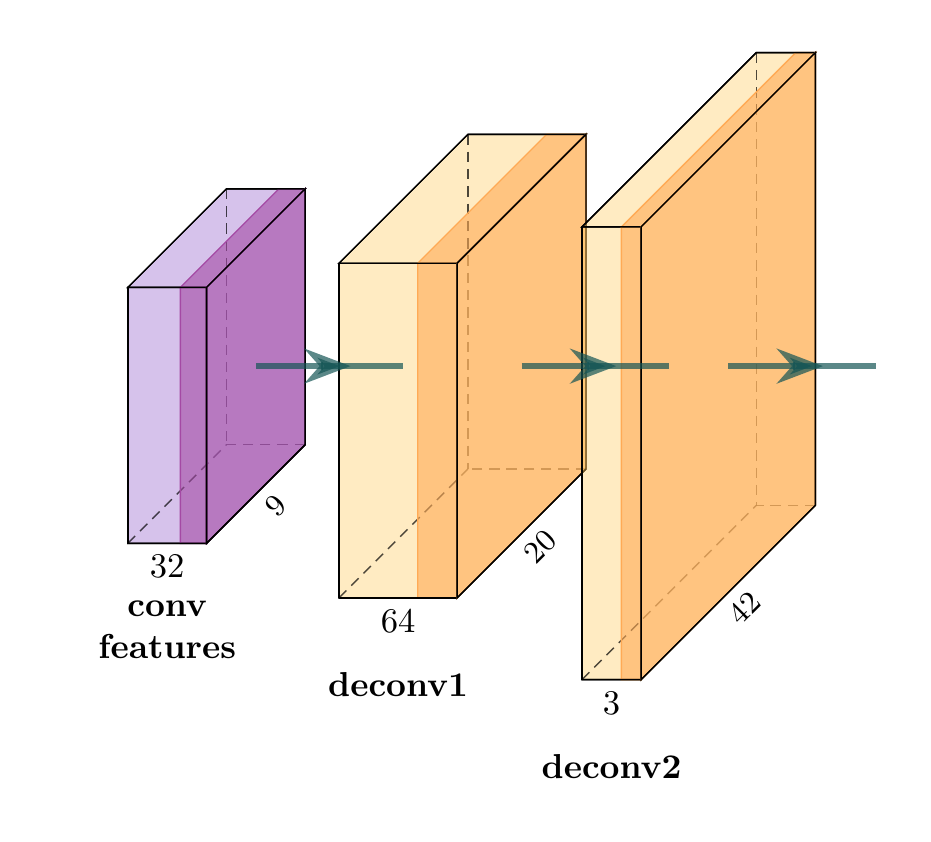}
    \caption[Image Segmentation Prediction Network]{Visual navigation auxiliary task network -- observation image segmentation and target segmentation prediction.}
    \label{fig:aux_head}
\end{figure}

Motivated by \cite{5596468} and \cite{mirowski2016learning}, we introduced additional auxiliary tasks that are specific to visual navigation. They were designed to enhance the training process as well as to help the network generalize. We train the model to predict the depth-map, image segmentation of the observation, and image segmentation of the target. For the image segmentations, we map the object-type to the RGB color space and maximize the distances between each color in the HSB color space. The input is passed through a narrow part of the network in autoencoder fashion to improve the quality of features in the shared part of the network. This gives the actor and the critic good features in bottom-most layers with a compact representation of all information needed to reconstruct depth-map and image segmentations. These bottom-most layers would otherwise be difficult to train since the network is deep and the loss is noisy due to the imprecise target values computed using the RL algorithm. The image segmentation for the target ensures the network pays attention to what the target is. Otherwise, it would be difficult for the network to take the target input into account at the beginning of the training.

For each visual navigation auxiliary task, there is a network attached to the last convolutional layer consisting of deconvolutional layers. The network architecture for the observation image segmentation and the target image segmentation can be seen in \figref{fig:aux_head}. For the depth-map prediction the structure of the network is the same, but the intermediate deconvolutional layer has only 32 filters, and the last layer has a single channel. The true features (the image segmentations for observation and the target and the depth map) are downsampled to a smaller size. The \gls{MSE} is computed between the outputs of the networks and the true features.

The additional auxiliary tasks for visual navigation also allow for the use of supervised learning to initialize the network with good features in the bottom-most part of the network since these are the least dependent on the policy. It is costly to render a 3D scene, but it is cheap to pre-compute a data set of observations taken from the scene and use it for supervised training.

\subsection{Environment Complexity}
The training of the agent might be hard, especially when the environment is large and the initial state is far from the target. To make the task easier for the agent, we first sample the initial states closer to the target and gradually increase the distance between the initial state and the target. Let $\tau \in [0,1]$ be the environment complexity. We define the maximal sampling distance $\dmax_E \colon [0,1] \rightarrow \mathbb{R}$ of an environment $E$ as follows:
\begin{equation}
    \dmax_E(\tau) = \tau \max_{s_1 , s_2}\{\dist(s_{1}, s_2) \} \, ,
\end{equation}
where $\dist(\cdot, \cdot)$ measures the distance between any two states of the given environment $E$. Any distance measure can be used, e.g., the Euclidean distance between the corresponding agent positions in the environment.

The initial state $s_0$ is sampled from a uniform probability distribution over the set of possible initial states closer to any target than $\dmax_E(\tau)$:
{
    \thinmuskip=\muexpr\thinmuskip*5/8\relax
    \medmuskip=\muexpr\medmuskip*5/8\relax
    \begin{equation}
        \mathbb{U}(\{ s_1 | s_1 \in \mathcal{S}_{start}, s_2 \in \mathcal{S}_{target}, \dist(s_1, s_2) \leq \dmax_E(\tau) \}) \, ,
    \end{equation}
}
where the set of target states is denoted by $\mathcal{S}_{target}$. The environment complexity $\tau$ starts at a low value, e.g. $0.3$, and gradually increases during the training to $1.0$.

\section{Experiments}
We have experimentally evaluated the performance of our method A2CAT-VN,  using the average episode length and the average episode undiscounted return as performance metrics. The averages are computed Monte Carlo estimates based on 100 rollouts. The randomness comes from the initial state, the non-deterministic behavior of the environment, and the stochasticity of the actor.

\subsection{Environments}

We have employed three different 3D environment simulators suitable for visual navigation tasks.

\begin{figure}[htbp]
    \centering
    \begin{tabular}{cc}
    \subcaptionbox{NavMaze 1}{\includegraphics[width=0.4\linewidth]{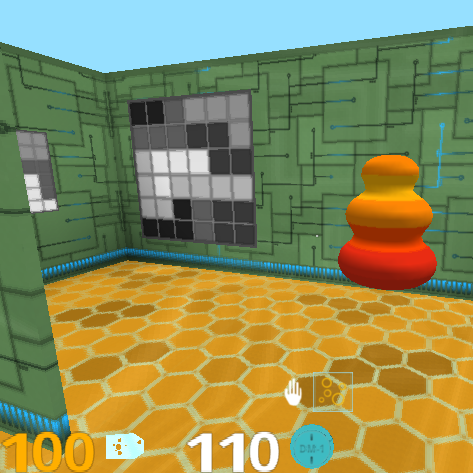}} &
    \subcaptionbox{NavMaze 1}{\includegraphics[width=0.4\linewidth]{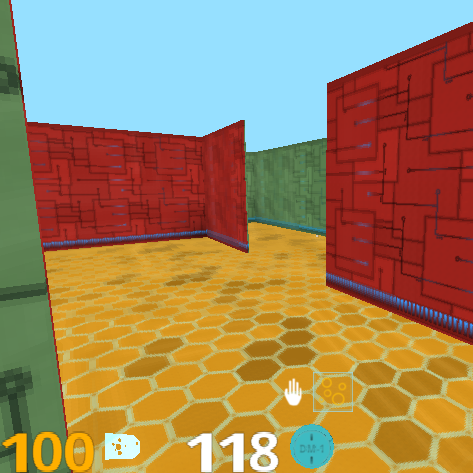}} \\
    \subcaptionbox{SeekAvoid}{\includegraphics[width=0.4\linewidth]{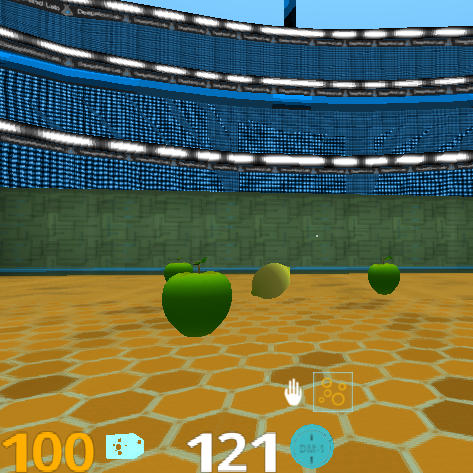}} &
    \subcaptionbox{NavMaze 2}{\includegraphics[width=0.4\linewidth]{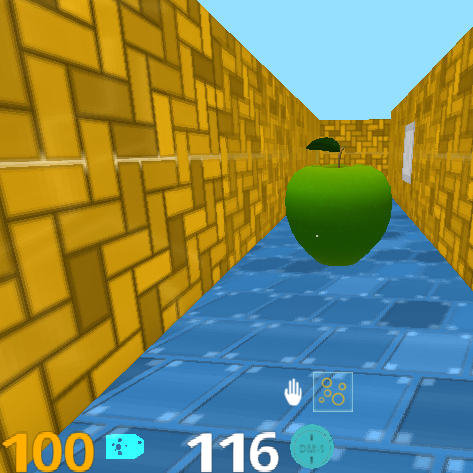}}
    \end{tabular}
    \caption[DeepMind Lab scene images]{Sample images from DeepMind Lab framework \cite{beattie2016deepmind}.}
    \label{fig:deep_mind_lab}
\end{figure}

\emph{1) DeepMind Lab} \cite{beattie2016deepmind} is a 3D framework which allows an agent to move and collect objects in synthetic environments. It is fast and highly optimized for training AI agents and the set of allowed actions is customizable. \figref{fig:deep_mind_lab} shows examples of images from this environment. We used it to compare the proposed algorithm with alternatives from the literature and to pre-train the agent's network for other environments, which sped up the training process.

\begin{figure}[htbp]
    \centering
    \begin{tabular}{cc}
    \subcaptionbox{bedroom\label{fig:bedroom}}{\includegraphics[width=0.4\linewidth]{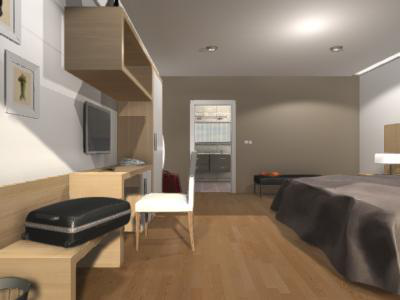}} &
    \subcaptionbox{bathroom\label{fig:bathroom}}{\includegraphics[width=0.4\linewidth]{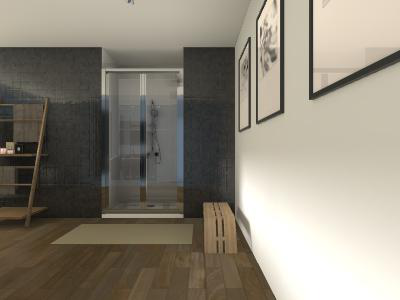}} \\
    \subcaptionbox{kitchen\label{fig:kitchen}}{\includegraphics[width=0.4\linewidth]{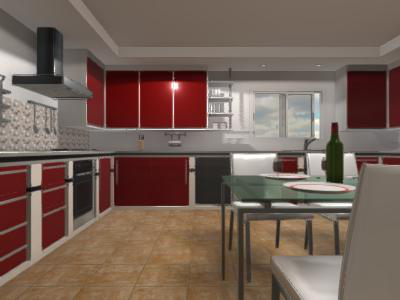}} &
    \subcaptionbox{living room\label{fig:living_room}}{\includegraphics[width=0.4\linewidth]{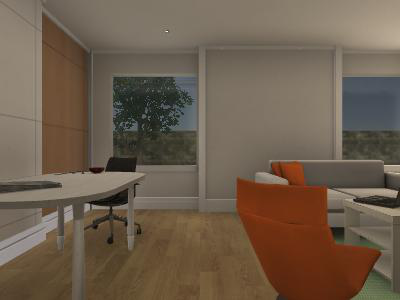}}
    \end{tabular}
    \caption[AI2-THOR scene images]{Sample images from AI2-THOR framework \cite{ai2thor}.}
    \label{fig:thor_cene_samples}
\end{figure}

\emph{2) AI2-THOR} \cite{ai2thor} is a photo-realistic interactive framework with high-quality indoor images (see \figref{fig:thor_cene_samples}). Most of the environments are a single room and  are dynamic, i.e., at the beginning of the episode, various objects can be placed at random positions. The agent moves on a grid: an action moves the agent to a neighboring point on the grid or rotates the agent by $90\,^\circ$. This does not allow for a good generalization since the agent can memorize the finite (and small) number of observations it can receive. Therefore, we have modified the implementation of the AI2-THOR 3D simulator to use continuous space. We have extended the set of possible actions by adding a rotation by an arbitrary angle and a movement by an arbitrary distance. We have also implemented the physics of collisions.

\begin{figure}[htbp]
    \centering
    \begin{tabular}{cc}
    \includegraphics[width=0.4\linewidth]{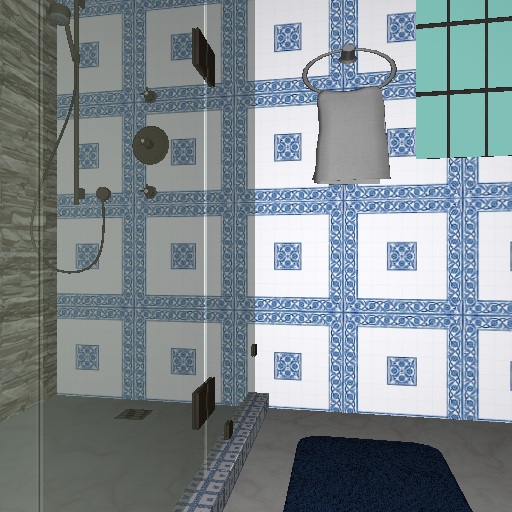} &
    \includegraphics[width=0.4\linewidth]{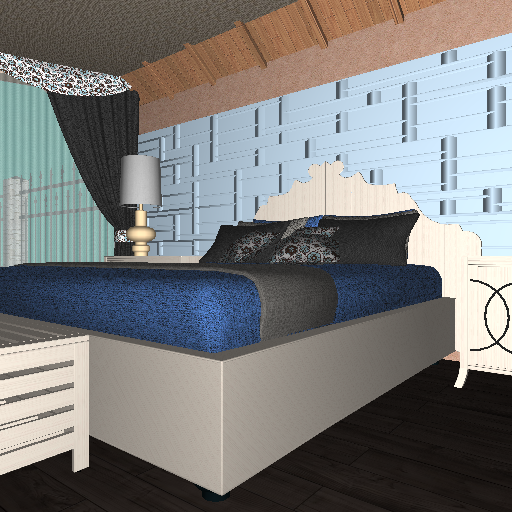} \\[0.25cm]
    \includegraphics[width=0.4\linewidth]{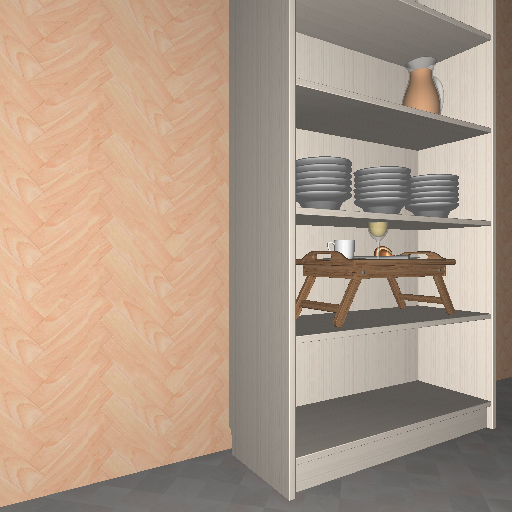} &
    \includegraphics[width=0.4\linewidth]{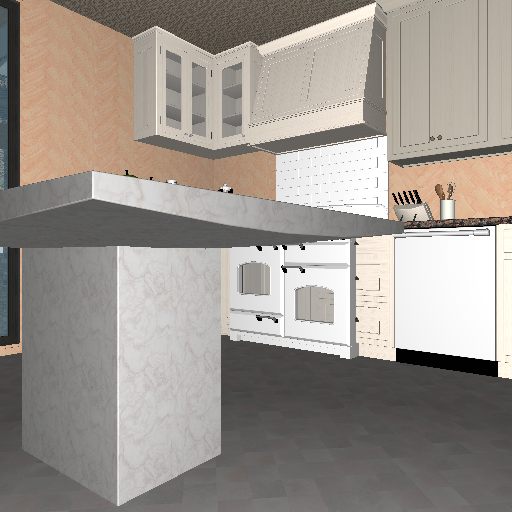}
    \end{tabular}
    \caption[SUNCG scene images]{SUNCG \cite{song2016ssc} scene images from the House3D \cite{wu2018building} framework.}
    \label{fig:house3d_scenes}
\end{figure}

\emph{3) House3D with SUNCG} \cite{wu2018building} is a 3D framework allowing to use the environments from the SUNCG dataset \cite{song2016ssc}. The SUNCG data set consists of over 45\,000 indoor environments, most of them being two-storey houses and studios. House3D is highly optimized for AI agents training and runs fast on GPUs. Apart from RGB output rendering, it also supports depth map and image segmentation rendering. Illustrative images from this environment are shown in \figref{fig:house3d_scenes}. The set of actions can be customized in a similar way as in the DeepMind Lab environment.
\subsection{Action Space}
\label{sec:actions}
In each of our experiments, we used actions from the following set: forward, backward, left, right, rotate-left, rotate-right, tilt-up, tilt-down. The forward and backward actions move the agent in the direction it is currently facing. The left and right actions move the agent in perpendicular directions to the direction it is facing. The rotate-left and rotate-right actions rotate the agent by $30$\,degrees\footnote{One experiment uses $90\,^\circ$ angles.} counter-clockwise and clockwise respectively and the tilt-up and tilt-down actions tilt the agent's camera up or down by $30$\,degrees.
%
%\begin{enumerate}
%    \item Forward -- moves the agent forward by $0.5$\,meters in the direction it is currently facing.
%    \item Backward -- moves the agent by $0.2$\,meters in the direction opposite to the direction it is currently facing.
%    \item Left -- moves the agent to the left by $0.35$\,meters, relative to the direction it is currently facing.
%    \item Right -- moves the agent to the right by $0.35$\,meters, relative to the direction it is currently facing.
%    \item Rotate-left -- rotates the agent by $30$\,degrees counter-clockwise.
%    \item Rotate-right -- rotates the agent by $30$\,degrees clockwise.
%    \item Tilt-up -- tilt the agent's camera up by $30$\,degrees.
%    \item Tilt-down -- tilt the agent's camera down by $30$\,degrees.
%\end{enumerate}
%

In real-world environments, the actuators would rarely be able to move the agent precisely. To simulate such a setting, Gaussian noise is added to the position and rotation of the agent after taking an action. More specifically, let $\tilde{s} = (\tilde{x_1}, \tilde{x_2}, \tilde{\theta}, \sigma)$ be the agent's position, horizontal rotation, and tilt of the camera after taking an action before we added the noise. Then the agent's final position and rotation is $s = (x_1, x_2, \theta, \sigma)$, with $x_1 \sim \mathcal{N}(\tilde{x_1}, 0.02^2)$, $x_2 \sim \mathcal{N}(\tilde{x_2}, 0.02^2)$ and $\theta \sim \mathcal{N}(\tilde{\theta}, 2^2)$.

\subsection{Training}
The reward can be assigned to the agent using different schemes. In our work, we give the agent a reward of one if it reaches the target and zeroes otherwise. In the training phase, we compute the total gradient as the weighted sum of all the partial gradients: the actor, the critic, the entropy loss, the off-policy critic, and the auxiliary tasks. The gradient is clipped, so its $l2$-norm does not exceed $0.5$ and the RMSprop optimizer is used to optimize the weights. In all experiments, we used two Tesla K40 GPUs (10GB each) -- one GPU was dedicated for the environments and the other one for the agent. The parameters used in our method are given in Table~\ref{tab:parameters}, where $f$ denotes the number of frames processed so far and $f_{\max}$ is the maximum number of frames to be processed during training. Some parameters were chosen to be the same as in \cite{jaderberg2016reinforcement,Wu2017ScalableTM}, others were tuned experimentally.
\begin{table}[htbp]
    \caption{Method parameters}
    \label{tab:parameters}
    \centering
    \begin{tabular}{lr}
        \toprule
        name & value \\
        \midrule
        discount factor ($\gamma$) & $0.99$ \\
        maximum episode length & 900 \\
        maximum rollout length & 20 \\
        maximum number of frames ($f_{\max}$) & $4 \cdot 10^7$ \\
        number of environment instances & 16 \\
        replay buffer size & $2\,000$ \\
        \midrule
        optimizer & RMSprop \\
        RMSprop alpha & $0.99$ \\
        RMSprop epsilon & $10^{-5}$ \\
        learning rate & $7 \cdot 10^{-4}\frac{f_{\max} - {f}}{f_{\max}}$ \\
        max. gradient norm & 0.5 \\
        \midrule
        entropy gradient weight & 0.001 \\
        actor weight & 1.0 \\
        critic weight & 0.5 \\
        off-policy critic weight & 1.0 \\
        pixel control weight & 0.05 \\
        reward prediction weight & 1.0 \\
        depth-map prediction weight & 0.1 \\
        observation image segmentation prediction weight & 0.1 \\
        target segmentation prediction weight & 0.1 \\
        pixel control discount factor & 0.9 \\
        pixel control downsize factor & 4 \\
        auxiliary VN downsize factor & 4 \\
        \midrule
        pre-training optimizer & Adam \\
        pre-training total epochs & 30\\
        pre-training dataset size & $2\cdot 10^5$ \\
        \bottomrule
    \end{tabular}
\end{table}

\subsection{Partial Observability}\label{sec:lstm_comp}

\begin{figure}[htbp]
    \centering
    \includegraphics[width=\linewidth]{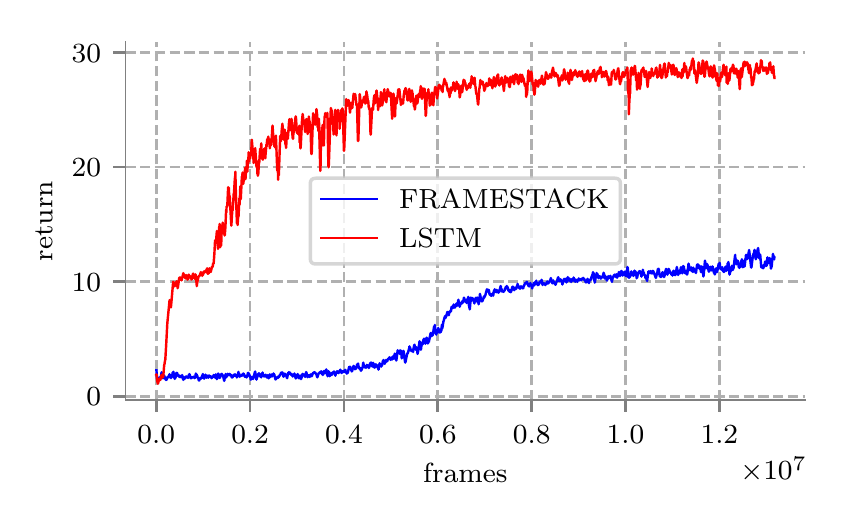}
    \caption[LSTM and frame-concatenation comparison]{Comparison of LSTM model (LSTM) with frame-concatenation model (FRAMESTACK) trained using deterministic UNREAL.
    The plot shows the average return curves during training on the DeepMind Lab \cite{beattie2016deepmind} environment called SeekAvoid.} %\ED{Improve caption (articles etc.)}
    \label{fig:a2c_lstm}
\end{figure}

We compared two different approaches to resolve the partial observability problem. One approach used by \cite{7989381, mnih2013playing} concatenates the past four frames as the input to the agent. The other approach \cite{jaderberg2016reinforcement} uses the LSTM network \cite{doi:10.1162/neco.1997.9.8.1735}. We tested both methods on the DeepMind Lab environment because of its great speed and relative simplicity. The allowed actions were forward, backward, left, right, rotate-left, rotate-right. We did not use any noise and the distance by which the actions moved the agent were $0.3$\,meters for actions forward, backward, left, right. The input to the agent was a single RGB image with the resolution of $84 \times 84$\,pixels. The network structure based on \cite{jaderberg2016reinforcement} was similar in both cases except in the frame concatenation version, where the LSTM was replaced by a linear layer. Both networks used the UNREAL auxiliary tasks \cite{jaderberg2016reinforcement}. The algorithms were trained on the DeepMind Lab SeekAvoid environment. The results can be seen in \figref{fig:a2c_lstm}. The experiment clearly shows that LSTM outperforms the frame concatenation method.

\subsection{AI2-THOR}
\begin{figure}[htbp]
    \centering
    \includegraphics[width=\linewidth]{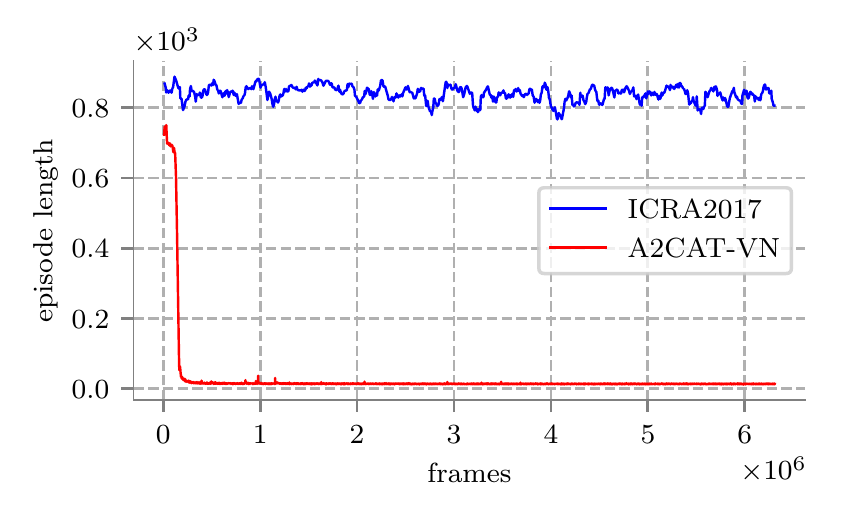}
    \caption[AI2-THOR baseline comparison]{Comparison of our model (A2CAT-VN) with target driven visual navigation paper \cite{7989381} (ICRA2017) trained on four environments from AI2-THOR.
    The plot shows the average episode length curves during training.}
    \label{fig:ai2thor_compare}
\end{figure}
We have trained our algorithm on four environments from AI2-THOR environment with multiple targets. We have used the same set of actions as \cite{7989381} -- rotate-left, rotate-right, forward, and backward. Actions forward and backward moved the agent in the direction is was facing by either $0.33$\,m or $-0.33$\,m. Actions rotate-left and rotate-right rotated the agent by $\pm 90^{\circ}$. No noise was applied. This allowed us to compare our method to \cite{7989381} and also to cache the observations since it turned the problem into an instance of a grid world. The resolution of the input images was $174\times 174$\,pixels. We used 16 environments in parallel for our algorithm each using a different scene or a different target.
We did not use any pre-training nor did we increase the environment complexity. Our method is compared to \cite{7989381}. The environments we chose for this experiment were bigger and more difficult to navigate than those used in \cite{7989381}, but came from the same AI2-THOR simulator. The training with our algorithm took roughly one day, while it took three days to train the network by using the algorithm described in \cite{7989381}. The results can be seen in \figref{fig:ai2thor_compare}. Our method A2CAT-VN found the optimal solution after approximately $5\cdot 10^5$\,frames, whereas the method described in \cite{7989381} was not able to converge even in $7 \cdot 10^6$\,steps.

\subsection{Continuous AI2-THOR}\label{sec:exp_cthor}
% maybe remove this experiment. Waiting for it to complete
\begin{figure}[htbp]
    \centering
    \includegraphics[width=\linewidth]{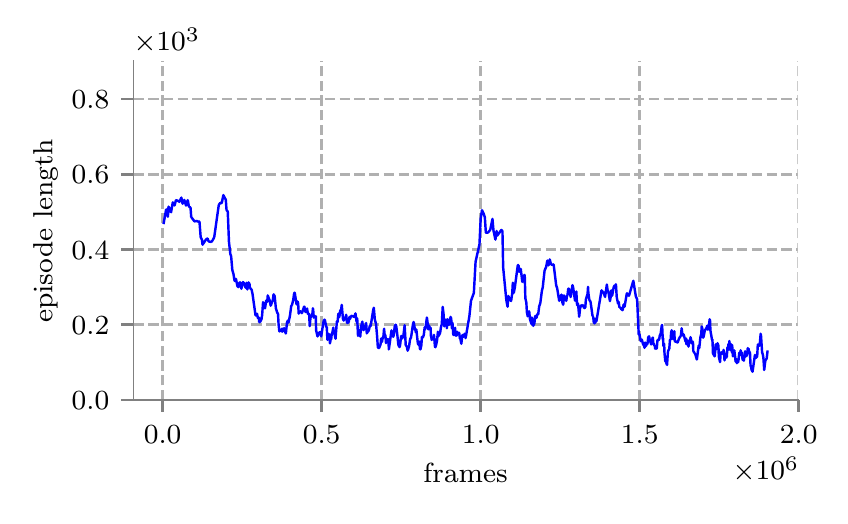}
    \caption[Continuous AI2Thor]{This plot shows the average episode length curve while training our algorithm A2CAT-VN in a single environment from AI2-THOR framework \cite{ai2thor} with multiple randomly placed targets.}
    \label{fig:ai2thor_multiscene}
\end{figure}
We trained our agent on the modified version of the AI2-THOR environment, using the full set of actions as described in Section~\ref{sec:actions}: forward, backward, left, right,
rotate-left, rotate-right, tilt-up, tilt-down. The forward and backward actions moved the agent by $0.5$ and $-0.2$ meters, respectively, and the left and right actions moved
the agent by $0.35$ meters. Due to performance issues, however, the noise was only applied in the direction of the movement and no noise was applied in case of tilt-up and tilt-down
actions. The agent was trained on a single bedroom scene with multiple targets specified by images. We used 16 environments in parallel, each having a different target.
The target object was placed randomly to different positions in the environments and the agent was trained to get to close proximity of the target object (1\,meter).
The resolution of the input images was $174 \times 174$\,pixels. We did not use pre-training nor did we increase the environment complexity. The results can be seen in
\figref{fig:ai2thor_multiscene}. The training took 4 days. The AI2-THOR 3D environment simulator was too slow for further experiments. The results show the ability of the agent to navigate in non-stationary environments and to recognize different objects in the scene.

\subsection{Auxiliary Tasks}
\begin{figure}[htbp]
    \centering
    \includegraphics[width=\linewidth]{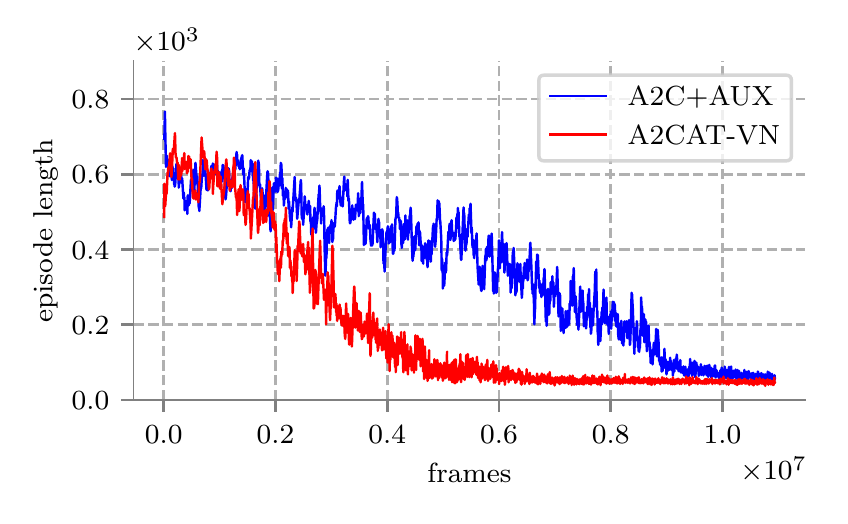}
    \caption[Auxiliary tasks comparison]{Comparison of our method (A2CAT-VN) with the A2C algorithm extended with UNREAL auxiliary tasks \cite{jaderberg2016reinforcement}.
    The training was performed on 16 environments from SUNCG dataset \cite{song2016ssc}.}
    \label{fig:auxiliary_tasks_comparison}
\end{figure}

We compared our method (A2CAT-VN) with the batched A2C extended with the original two UNREAL auxiliary tasks. A single agent was trained on 16 houses chosen randomly from a subset of the SUNCG dataset \cite{song2016ssc} using House3D environment simulator. We used the same actions as those described in Section~\ref{sec:exp_cthor} except for the tilt-up and tilt-down actions. Inspired by \cite{wu2018building}, the agent was trained to find a selected room in the house. The room was given to the agent in the form of an observation taken in a room of the same type. For example, if the target room is a bedroom, the agent is supposed to find any bedroom. The resolution of the input  images was $174 \times 174$\,pixels. We pre-trained our neural network using the data collected from a subset of all houses from SUNCG dataset. The number of images we used for pre-training was approximately $20\,000$ and we trained our network for 30 epochs using the Adam optimizer. For the full training, we linearly increased the environment complexity from 0.3 at time step 5M to 1.0 at time step 10M. The training took roughly two days. The training curves for the average episode length can be seen in \figref{fig:auxiliary_tasks_comparison}. Our algorithm A2CAT-VN converged much faster with the additional auxiliary tasks for visual navigation enabled, reaching the average episode length of 200 in $\approx 3 \cdot 10^6$ frames whereas without the additional tasks the training took $\approx 8 \cdot 10^6$ frames to get to the same level.

\section{Conclusions \& Future Work}
We have proposed a novel learning architecture A2CAT-VN for visual navigation in indoor environments. It is based on a compact deep neural network capable of fast learning over multiple realistic environments, using the batched A2C algorithm extended with novel auxiliary tasks. By using the target image as an input, our method enables the agent to locate arbitrary goals, as long as their images have been used during the training phase.

The method was demonstrated on AI2-THOR and House3D environments. First, we have shown that the basic batched A2C algorithm benefits from the addition of the UNREAL auxiliary tasks \cite{jaderberg2016reinforcement}. A further performance gain was achieved by employing additional auxiliary tasks designed specifically for visual navigation.

When applied to AI2-THOR environment, our method was able to converge at least an order of magnitude faster than an alternative state-of-the-art method \cite{7989381}, which also allowed for the use of multiple targets and was demonstrated in indoor environments, similarly to our method. The auxiliary tasks introduced were shown to reduce the number of frames needed to train the agent by a factor of two and they allowed to use supervised learning to pre-train a part of the network.

Future research can focus on the potential effect of using supervised pre-training of additional auxiliary tasks for visual navigation on the training performance as well as the effects of individual additional auxiliary tasks through an ablation study. We would also like to explore the application of our method to more 3D environments (perhaps outdoor environments) and potentially apply it to mobile robots moving in real-world environments. Another line of research needs to be conducted on the ability of the method to generalize to unseen targets. In addition, we believe the ability of the agent to deal with unseen environments might outline an important area for future research.

%Our code was implemented in PyTorch \cite{paszke2017automatic}. The implementation of batched A2C with auxiliary tasks can be found is in repository \url{https://github.com/jkulhanek/deep-rl-pytorch}. The code needed to reproduce our result is in repository \url{https://github.com/jkulhanek/target-driven-visual-navigation.git}.

%\section{Acknowledgements}

\bibliographystyle{IEEEtran}
\balance
\bibliography{IEEEabrv,ecmr2019visnavrl}

\end{document}